\documentclass[letterpaper, 10 pt, conference]{ieeeconf}  

\IEEEoverridecommandlockouts                              
\usepackage{amsmath}
\usepackage{amssymb}
\usepackage{amsthm}
\usepackage[backend=bibtex,style=ieee]{biblatex}
\usepackage{algorithm}
\usepackage[noend]{algpseudocode}

\usepackage{graphicx}
\usepackage{caption}
\usepackage{subcaption}
\usepackage{xcolor}

\newtheorem{proposition}{Proposition}

\newcommand{\Ac}{\mathcal{A}}

\newcommand{\Ec}{\mathcal{E}}

\newcommand{\Mc}{\mathcal{M}}

\newcommand{\Yc}{\mathcal{Y}}

\newcommand{\E}{\mathbb{E}}

\newcommand{\rdnote}[1]%
    {\textcolor{blue}{[RD: #1]}}
\newcommand{\emnote}[1]%
    {\textcolor{orange}{[EM: #1]}}
\newcommand{\vrnote}[1]%
    {\textcolor{purple}{[VR: #1]}}  

\title{\LARGE \bf
Expert Selection in High-Dimensional Markov Decision Processes
}

\author{
  Vicen\c{c} Rubies-Royo$^*$, 
  Eric Mazumdar$^*$,
  Roy Dong$^*$,
  Claire Tomlin,
  and S. Shankar Sastry\\
\thanks{The authors are with the department of Electrical Engineering and Computer Science, University of California, Berkeley, Berkeley, CA. $^*$ indicates equal contribution.}}

\begin{document}

\maketitle
\thispagestyle{empty}
\pagestyle{empty}



\begin{abstract}
In this work we present a multi-armed bandit framework for online expert selection in Markov decision processes and demonstrate its use in high-dimensional settings. Our method takes a set of candidate expert policies and switches between them to rapidly identify the best performing expert using a variant of the classical upper confidence bound algorithm, thus ensuring low regret in the overall performance of the system. This is useful in applications where several expert policies may be available, and one needs to be selected at run-time for the underlying environment.
\end{abstract}


\section{Introduction}
\label{sec:intro}

Markov decision processes (MDPs) represent a mathematical framework for dealing with decision problems in many fields. It is usually hard, however, to predict how changes in the underlying MDP might affect the performance of a given policy, especially if changes happen online. For example, when the observation model is not accurate due to corrupt/noisy observations from damaged sensors, or the dynamics model becomes suddenly inaccurate due to underlying changes of the system, it is challenging to know whether the current policy will perform in a satisfactory manner. Trying to re-learn a policy online using reinforcement learning may be an option, even though the large amount of data required to update the parameters of the model might be prohibitive \cite{Minh_2013,Schulman_15_1,Schulman_15_2}. Other ways to circumvent this problem may be to use noise at training time \cite{basu2017learning,Dodge2016} or use some filtering procedure \cite{duan2010highly,jassim2013image}. In this work we address this problem through the use of multi-armed bandits.

The multi-armed bandit problem represents one the most simple settings to study the exploration vs. exploitation trade-off. The literature on multi-armed bandits is rich and contains a myriad of strategies for dealing with the problem, along with guarantees of performance in the form of regret bounds \cite{lai1985,Robbins1952,abbasi2011,Bianchi2012,jin2018qlearning,TorCsabaBandits}. In this work, we show that the task of choosing among a set of expert policies during the execution of a MDP, can be likened to a multi-armed bandit problem. In contrast to classical multi-armed bandits, there is strong coupling between the states and the rewards of the MDP, so one cannot freely `switch' between expert policies without repercussions. 

\begin{figure}[tbp]
  \centering
  \includegraphics[width=0.35\textwidth]{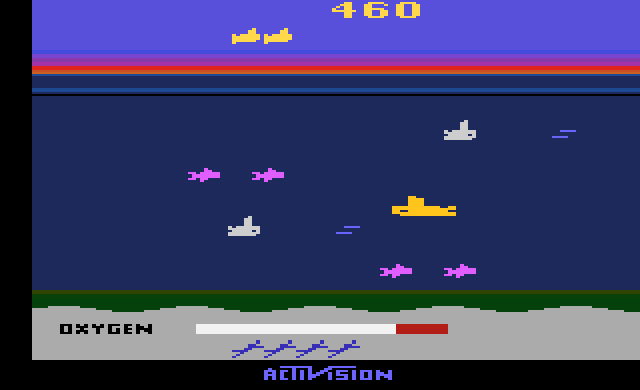}
  \caption{Snapshot of the Seaquest domain.}
  \label{fig:games_snapshot}
\end{figure}

The primary focus for this paper will be to present the experimental results of the algorithm on a low-dimensional and a high-dimensional MDP setting, while deferring most of the mathematical details to \cite{Mazumdar2017}. For the low-dimensional scenario, we will use a gridworld environment where different agents are trained under various dynamics models to reach a set of high-reward states. For the high-dimensional case, we will use the Seaquest video game environment, shown in Fig. \ref{fig:games_snapshot}, where each expert policy is trained under different observation models. Unlike the gridworld, agents trained in this environment receive state information through raw images from the game. Despite the high-dimensional observations, the multi-armed bandit algorithm is capable of selecting the best expert from our set at run-time to minimize regret and outperform experts specifically trained to be robust.

\section{Background}
\label{sec:back}


The multi-armed bandit problem was introduced in \cite{Robbins1952}, and has been an active field of research since. In \cite{lai1985} the classical Upper Confidence Bound (UCB) algorithm was introduced as a solution with logarithmically growing regret. More recently, multi-armed bandit (MAB) algorithms and the notion of regret have been used for the control of systems with dynamics. For example, in \cite{Kakade2004} a MAB formulation was used to train reinforcement learning agents, and \cite{Auer2009} constructed bounds on the regret of RL algorithms. Many other algorithms have guarantees on the growth of the regret; for an overview of some of the work see \cite{abbasi2011,Bianchi2012}.

In \cite{Cesa-Bianchi2006}, an overview is given of various bandit approaches for choosing an expert from a set of experts online under many different assumptions on the underlying structure. The idea of using pre-trained experts for control has also been investigated in various settings. \cite{Hamrick2017} used the notion of experts for meta-control of MDPs when there is a fixed computational cost and \cite{Laskey2015} used experts in conjunction with recommendation algorithms for choosing which actions to take in an uncertain environment.

The idea of expert selection in MDPs is also similar to the idea of restless bandits, where the underlying distribution of each arm evolves according to separate Markov processes that are independent of the pulls \cite{Bubeck2012}. Recent work has been conducted in deriving bounds on the regret of varying algorithms including the UCB algorithm in this setting \cite{Ortner2012,Tekin2012}. The critical difference in this case lies in the fact that in our setting, each pull of an arm (equivalently choice of an expert) effects the underlying dynamics which are also shared across all arms. Another related line of work seeks to use multi-armed bandit-inspired algorithms for reinforcement learning problems, such as the Q-Learning UCB approach analyzed in \cite{jin2018qlearning}. 

Thus, while MAB approaches for expert selection in MDPs tend to focus on isolated expert executions as the metric of performance for each bandit, in this work we define a sequence of time horizons $T_0,T_1,...$ and equate the action of pulling the arm of one of the bandits, to executing a specific expert on the MDP for $T_n$ steps. In our work we devise a version of the UCB algorithm and show that this online expert selection paradigm, despite violating the assumption of independence between successive pulls, can be seen as an ``almost bandit" problem. We then proceed to test the algorithm on a low-dimensional gridworld environment and a high-dimensional video game environment. 



\section{Problem Formulation and Analysis}
\label{sec:problem}

In our formulation, the controller is an algorithm that selects an expert policy to execute during the evolution of the underlying MDP. The chosen expert is fed observations of the state of the system, and outputs a distribution over the actions to take. The controller then samples the action, takes the action, and receives a reward from the environment. 

The controller is agnostic to the true state of the environment as well as the underlying dynamics. The \textit{only} information available to the controller is the history of rewards it has received for each expert. This problem lends itself to a multi-armed bandit formulation, although, as we will shortly see, it is significantly different from a classical multi-armed bandit problem. 

In this section, we introduce an Upper Confidence Bound (UCB) algorithm applied to expert selection in MDPs. We then provide some intuition for the algorithm and discuss its performance in terms of regret.

\subsection{Setup}

Before discussing the particular algorithm used to solve the problem, we introduce some notation. The environment of interest is modeled as an MDP, $\Mc$ which is represented as a 6-tuple $\Mc=(\Sigma,\Ac,P_a,R,\gamma,\mu_0)$,  where $\Sigma$ denotes the state space, $\Ac$, the action space, $P_a$ the state transition kernel, $R$, the reward kernel, $\gamma$ the discount factor, and $\mu_0$ the initial state distribution. Note that $P_a(\cdot,\cdot): \Sigma \rightarrow \Delta(\Sigma)$ for each action $a \in \Ac$, and $R(\cdot,\cdot,\cdot): \Sigma \times \Ac \times \Sigma \rightarrow \Delta([0,1])$, where $\Delta(X)$ denotes the set of all probability distributions over the set $X$. Further, we assume that there is an unknown observation kernel $P_y(\cdot): \Sigma \rightarrow \Delta(\Yc)$ that, for a given state $s$, returns a probability distribution over the observation space $\Yc$. Lastly, we are given a set of experts $\Ec=\{\pi_1,\pi_2,...\}$, where $\pi_i: \Yc \rightarrow \Delta(\Ac)$.

\subsection{Upper confidence bounds in MDPs}

In this section we provide the upper confidence bounds for expert selection in MDPs. For further details regarding the derivations we direct the reader to \cite{Mazumdar2017}. 

At each round $n=0,1,...,N$ the controller chooses an expert $e_n \in \Ec$ and follows its policy for an episode of length $T_n$. The controller then observes the average reward over the episode. Letting $t_{n}=\sum_{i=0}^{n-1}T_i$, we define the average reward for expert $e_n$ over the $n$-th episode as the random variable given by:
\begin{align}
r_{e_n}=\frac{1}{T_n}\sum_{t=t_{n}}^{t_{n}+T_n} R(s_t,a_t,s_{t+1}),
\end{align}
where $a_t\sim \pi_{e_n}(s_t)$. The controller keeps track of the average $r_{i}$ across all the times an expert $i$ has been chosen, which for simplicity we denote as $R_{i}$. Thus, if the algorithm has been run for $n$ rounds and expert $i$ has been chosen $n_i$ times, $R_i$ is given by
\begin{align}
R_i=\frac{1}{n_i}\sum_{n=0}^N r_{e_n}\mathbb{I}(e_n=i),    
\end{align}
where $\mathbb{I}$ is the indicator function. After having chosen an expert $i$, $n_i$ times, $R_i$ can be shown to be close to 
\begin{align}
\bar{R}_i =\lim_{m\rightarrow \infty}{\frac{1}{m}\E_{\pi_i}[\sum_{t=0}^m R(s_t,a_t,s_{t+1})| s_0 \sim \mu_0]},    
\end{align}
which represents the average reward under the stationary distribution for expert $i$. Given these definitions, the following relation holds between $\bar{R}_i$ and $R_i$:
\begin{align}
\Pr\left( \bar R_i \leq R_i + \sqrt{\frac{2}{n_i}\log{\frac{1}{\delta}}}+\frac{\max_{e \in \Ec} K_e}{T_0} \right) \geq 1-\delta.    
\end{align}
Here, the constant $K_e$ depends on the expert's policy and the underlying dynamics of the MDP, and $\delta$ is a chosen parameter. 
We note that $\frac{\max_{e \in \Ec} K_e}{T_0}$  adds a constant offset to each experts confidence bound, and as such, this term can be ignored when choosing between confidence bounds. We introduce the following notation to denote this approximate confidence bound for expert $i$:
\begin{align}
c_{i}=\sqrt{\frac{2}{n_i}\log{\frac{1}{\delta}}}.
\end{align}

When we implement our algorithm, this upper confidence bound will be treated as if it were the expected reward; this gives rise to an optimism-based approach.

\subsection{Modified upper confidence bound algorithm}

The classic UCB algorithm was introduced in \cite{lai1985} as a solution to the famous multi-armed bandit problem. Here we introduce the UCB algorithm variant for expert selection in MDPs.
Algorithm \ref{alg:ucb} is initialized with the confidence level $\delta$, a sequence of episode lengths $T_0,T_1,...$, and a set of expert policies $\Ec=\{\pi_1,\pi_2,...\}$. The expert policies are pre-trained and considered to be optimized for different settings of the environment. We note, that all policies must map from the same observation space $\Yc$ to the action space $\Ac$ of the MDP. These policies can be optimized for different noise conditions (modeled as different observation kernels), different tasks (modeled as different reward kernels), or even different dynamics (modeled as different state transition kernels). 

\begin{algorithm}[]
  \caption{Upper Confidence Bounds for MDPs}
  \begin{algorithmic}[1]
  	\State \textbf{Input:} $\delta$, pre-trained set of experts $\Ec$, an MDP $\Mc$, and a  episode lengths $T_0,T_1,...$
  	\State \textit{Initialize MDP:} $s_0 \sim \mu_0$ 
  	\State $t \gets 0$
  	\State $n_i \gets 0 \ \ \forall \ i \in \{1,...,|\Ec|\}$ \(\triangleright\){ number of expert calls.}
 	\State $c_{i} \gets \infty \ \ \forall \ i \in \{1,...,|\Ec|\}$
 	\State $R_i \gets 0 \ \ \forall \ i \in \{1,...,|\Ec|\}$ \(\triangleright\){ cumulative $T$-step rewards.}
 	\State $S_i \gets 0 \ \ \forall \ i \in \{1,...,|\Ec|\}$
 	\For{$n=0,1,2,...,N$}
		\State $e \gets \arg\max_{i} \{R_i+c_{i}\}$ \(\triangleright\){ ties broken arbitrarily.} \label{alg:argmaxline}
		\State $r\gets0$
		\For{$k=t$ to $t+T_n$}
			\State Receive observation $y_k$
			\State $a_k \gets \arg\max_{a \in \Ac} \pi_e(y_k)$
			\State Apply $a_k$ to environment, receive reward $r_k$
			\State $r\gets r+r_k$
		\EndFor
		\State \textbf{end for}
		\State $S_e \gets S_e+r/T_n$
		\State $n_e \gets n_e +1$
		\State $R_e \gets S_e/n_e$
		\State $t \gets t+T_n$
		\State Update $c_e$
 	\EndFor

	\end{algorithmic}
	\label{alg:ucb}
\end{algorithm}

We begin by initializing the confidence bound of each expert to infinity to ensure that each expert is chosen at least once. The algorithm works by choosing the expert policy for which it believes it can receive the highest steady-state reward, which as per the optimism-based approach will be the expert with the highest upper confidence bound. At each iteration $n$, the algorithm chooses an expert policy to execute for $T_n$ time steps. It collects the reward across the $T_n$ steps, updates its estimate of the expert's steady-state reward, and then updates the confidence bound of the expert, $c_{i}$ accordingly. The confidence bound $c_{i}$ is the maximum distance that the estimate can be from the true value of the average steady-state reward of the expert. This  bound holds with probability at least $1-\delta$. Thus in practice, $\delta$ is often chosen to be small.

As the algorithm tries out different experts and gathers more data, it tightens the confidence bound. By choosing the expert that has the maximum upper confidence-bound (Line~\ref{alg:argmaxline} in Algorithm~\ref{alg:ucb}), we get an optimism-based algorithm that picks the best guess at each time-step. The rationale for an optimism-based approach is that if the expert is suboptimal, the algorithm will eventually pick a different expert whose upper bound is better. 


 In contrast to the classical literature on multi-armed bandits, when we switch between experts, there is a strong coupling in the trajectory of the MDP: when an expert first takes control, the current state of the system is a consequence of actions taken by the previous (possibly different) expert. As shown in \cite{Mazumdar2017}, in this framework, if an expert is used for a sufficiently long period of time, ergodicity implies that the average reward seen will be close to the average steady-state reward of that expert. This is made explicit in the following corollary, which follows from basic properties of finite Markov chains.

\begin{proposition}
\label{corr:mc_bnd}
Suppose the induced Markov chains for each expert $e \in \Ec$ are irreducible and aperiodic. Pick any $n=0,1,...$, and let $e = e_{n}$ be the expert chosen at the $n$th iteration. We have that:
\begin{align}
\left| \bar R_{e} - \E\left[ \frac{1}{T_n}\sum_{t = t_{n}}^{t_{n}+T_n}R(s_t,a_t,s_{t+1}) \middle| s_{t_n},e \right] \right| \leq \frac{K_e}{T_n} \text{ a.s.}
\end{align}

\end{proposition}

We remark that $K_e$ is a constant that depends on the second largest eigenvalue of the induced Markov matrix of the expert, but not on the dimension of the state or action space. Given this property, it can be shown that this UCB-based algorithm for expert selection enjoys strong theoretical performance properties measured in terms of regret, which we define next.

In order to gauge the efficacy of the UCB algorithm it is necessary to introduce the idea of regret. We first point out that for regret to be practical, we need to calculate it with respect to a reasonable baseline. Let $\bar{R}_*=\max_{e \in \{1,2,...,N\}}{\bar{R}_e}$. Thus $\bar{R}_*$ is the best steady state average expected reward among all of our experts. We define the regret of the algorithm until iteration $n$ to be
\begin{align}
r(n)=n\bar{R}_*-\sum_{k=0}^{n}\frac{1}{T_k}\sum_{t=t_k}^{t_{k}+T_k}R(s_t,a_t,s_{t+1}),
\end{align}
where $t_{k}$, as before is given by $t_{k}=\sum_{i=0}^{k-1} T_k$. By virtue of corollary \ref{corr:mc_bnd}, algorithm \ref{alg:ucb} has the property that its regret is logarithmic in the time horizon for a properly chosen sequence length of episodes $T_0,...,T_n$ such that $T_n\ge T_0$ for all $n\ge0$. Indeed, let us define the expected regret of expert $e$ as $\Delta_e = \bar R_* - \bar R_e$. Note that $\Delta_{e^*} = 0$. Then it can be shown that the UCB algorithm for expert selection in MDPs has the following regret guarantee:

\begin{proposition}[Regret bound for algorithm \ref{alg:ucb}]
\label{prop:ucbregret}
~\\For a set of experts $\Ec$ and initial time horizon $T_0$ such that ${\bar{R^*}-\bar{R}_e > \frac{2K_e}{T_0}}$ for all $e \in \Ec$, and for a choice of ${\delta(n)=n^{-4}}$, the expected regret of the UCB algorithm after $n$ iteration steps, $\E[r(n)]$, satisfies:
\begin{align}
\begin{split}
\E[r(n)]&\le \sum_{\substack{e \in \Ec\\ e\ne e^*}}{\bigg[ \bigg( \frac{32 \log{n}}{(\Delta_e -\frac{2K_e}{T_n})^2} + c_1 \bigg) c_e}\bigg] + \sum_{k = 0}^{n-1} \frac{K_*}{T_k},
\end{split}
\end{align}
\end{proposition}
\noindent where $c_1 = 1+\pi^2/3$ and $c_e = \Delta_e+\frac{K_e}{T_0}$. For brevity we direct the reader to \cite{Mazumdar2017} for the proof of the above proposition which follows from standard proof techniques in the multi-armed bandits literature and makes use of corollary~\ref{corr:mc_bnd} to bound the bias between the finite-time approximation to the steady-state reward of an expert. 

\begin{figure}[tbp]
 \centering
 \includegraphics[width=0.75\columnwidth]{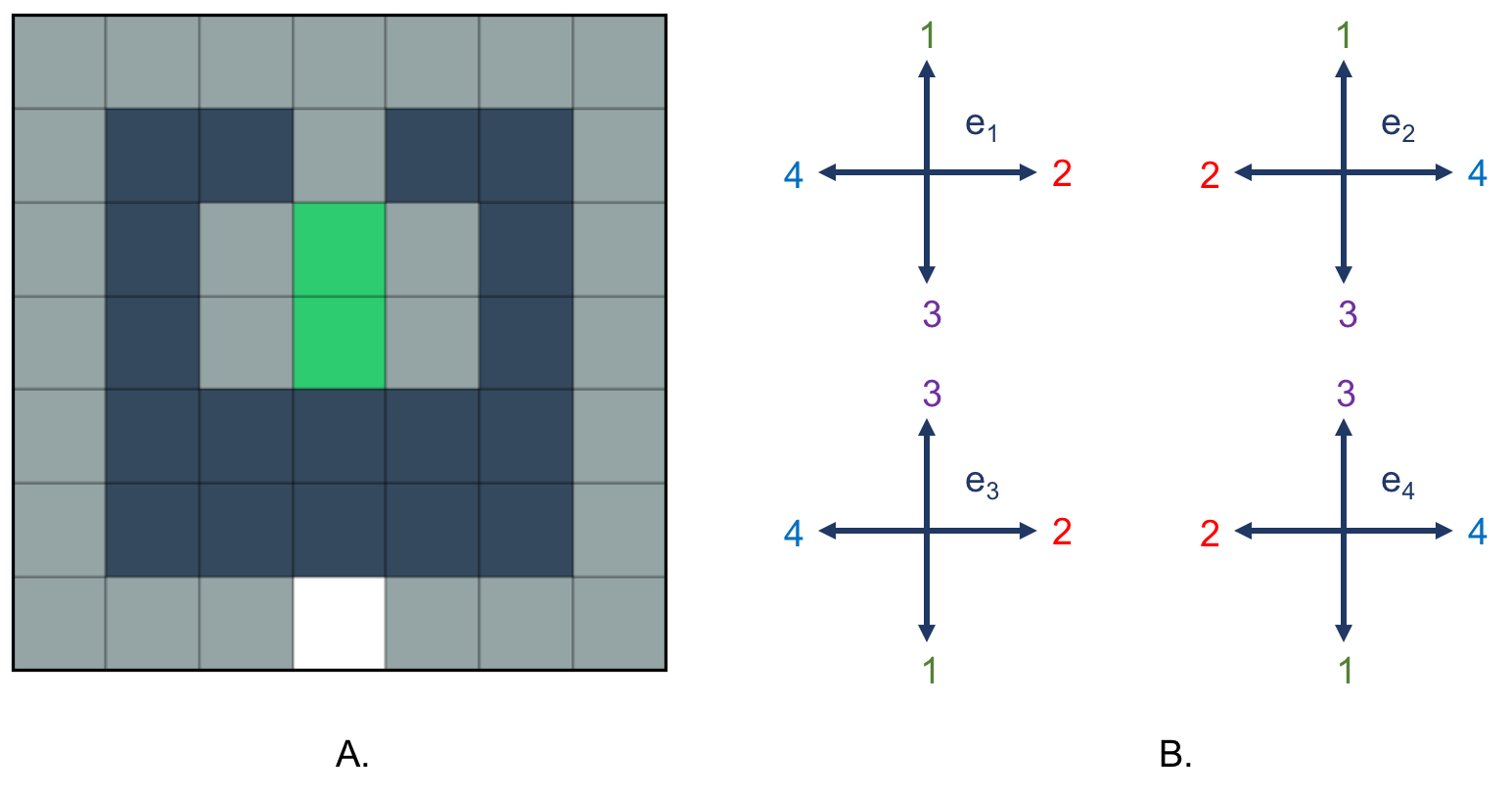}
 \caption{A. The setup of the Gridworld used for the experiments. B. The dynamics under which each expert $e_1,e_2,e_3,e_4$ were trained.} 
\label{fig:setup}
\end{figure}


\section{Experiments}
\label{sec:application}


In this section we present two numerical experiments to validate the proposed method. The first is a simple low-dimensional gridworld where each expert was trained with a different dynamics model, and the second is a game environment with an extremely high-dimensional obervation space where the experts were trained under different observation kernels in the form of occlusions.

\subsection{Low-dimensional Gridworld}

We first test the UCB algorithm outlined in Section \ref{sec:problem} on a gridworld example. The goal of the ``agent" in the gridworld MDP is to maximize the cumulative reward by reaching a set of goal states. We first describe the dynamics, and then the reward structure of the grid.

The grid and setup used are shown and described in Fig. \ref{fig:setup}A. Each state, in this MDP, is a tile of the grid, and the agent begins in the white tile. The action space is $\Ac=\{1,2,3,4\}$. Each action corresponds to a move in one of the four cardinal directions. The dark blue states are almost-trapping states, where with probability 0.98 the agent is stuck in that state, and with probability 0.02, the agent moves in the direction of their chosen action. For all other states, the agent follows its desired action with probability 0.97, and goes in a random other direction with probability 0.03. In states along the edge of the grid, an action going out of the grid will result in a movement in a random direction back into the grid. 

The rewards in the grid are as follows: the green squares both give a reward of 1, while the gray squares give rewards of 0.1. Dark blue squares give a reward of 0. Given this setup, the optimal policy would have the agent move around the edge of the grid, avoiding the dark-blue states, and then go back and forth between the two green states.

The agent, in this problem, is supplied with a collection of expert policies $e_1,e_2,e_3,e_4$, and no knowledge of the underlying true dynamics. Each expert policy was trained on the same MDP structure but with different dynamics models. For example, for experts $e_1$ and $e_2$, the action $1$ corresponds to a movement `up' in the grid, whereas for experts $e_3$ and $e_4$, action $3$ is `up'. The map from actions to movement that each expert was trained under is shown in Fig. \ref{fig:setup}B.

Given this grid-world  MDP and collection of experts $\Ec=\{e_1,e_2,e_3,e_4\}$, we now run the UCB algorithm with  episode lengths $T_n=\lceil T_0+cn\rceil$, where we set $c=0.1$ for all the experiments and keep $T_0$ as a variable for now. 


\begin{figure}[t]
\centering
\begin{subfigure}[b]{0.5\columnwidth}
  \centering
  \includegraphics[width=\textwidth]{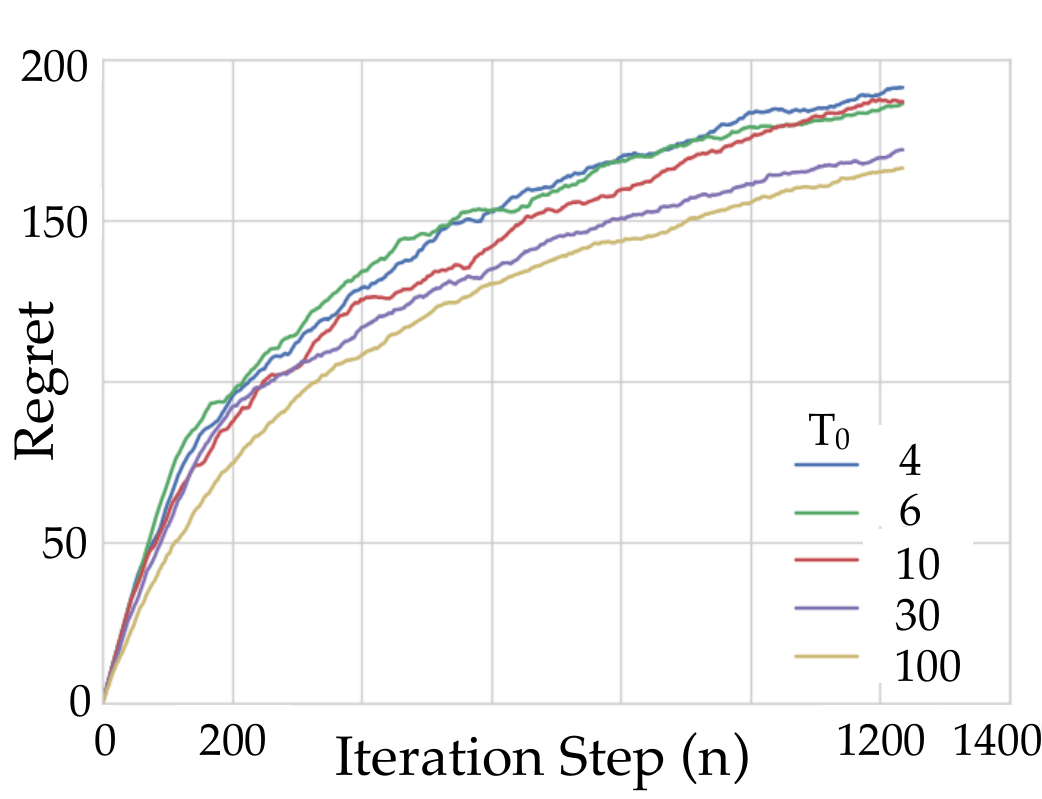}
  \caption{Average regret.}
  \label{fig:regret}
\end{subfigure}%
\hfill
\begin{subfigure}[b]{0.5\columnwidth}
  \centering
  \includegraphics[width=\textwidth]{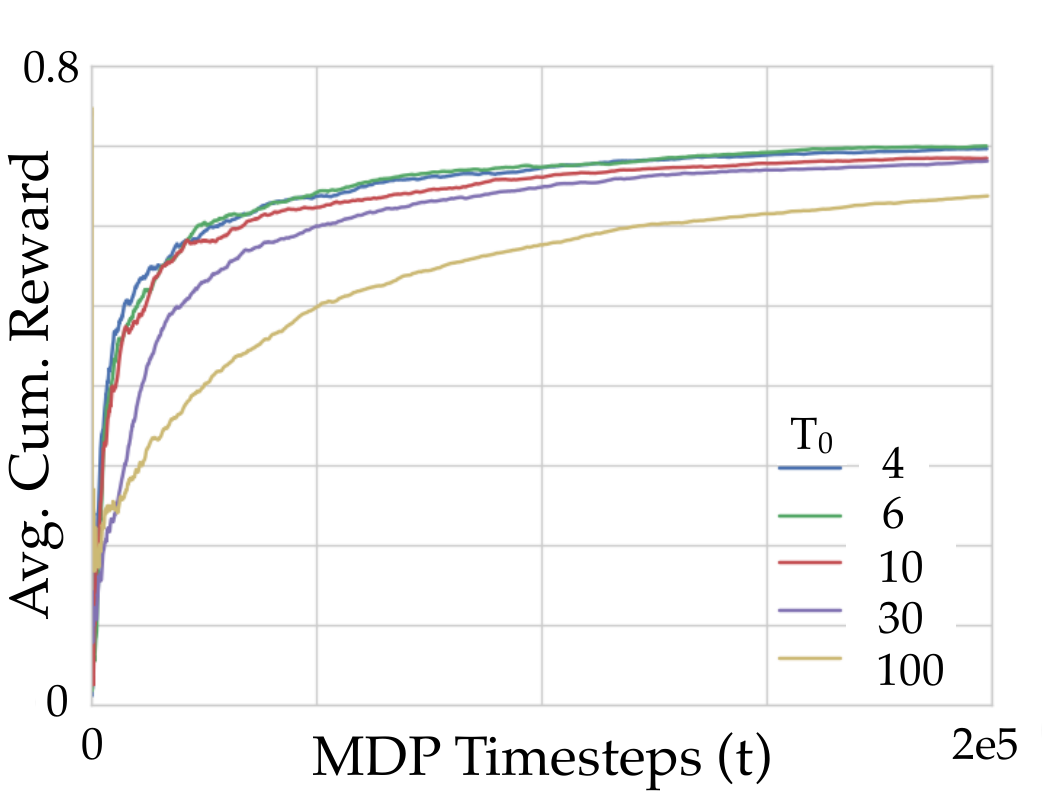}
  \caption{Average cumulative reward.}
  \label{fig:reward}
\end{subfigure}
\caption{Results from running the UCB algorithm 10 times for various values of $T_0$ and then taking the average.} 
\label{fig:results}
\end{figure}

We note that the MDP is simple by construction, and we use it more for illustrative purposes than to show the algorithm at work on a difficult task. Further, we note that the confidence bounds and sequence of time horizons we use for this experiment are those used in the derivation of the regret bound in \cite{Mazumdar2017}. There are choices that can possibly yield better qualitative results, but for the sake of consistency we stay with the same choices as in the derivation.

Given the setup in the prequel, we now show the expected regret of the UCB algorithm. We run the algorithm 10 times for various values of $T_0$ and plot the average regret in Fig. \ref{fig:regret}. The values of $K_e$, $\bar{R}_e$, and $\Delta_e$ for each expert in $\Ec$ can be found in \cite{Mazumdar2017}. We note that expert $e_1$ is the best expert.




We can clearly see in Fig. \ref{fig:regret}, the logarithmic rate of growth of the regret. Further, we can see how long time horizons lead to lower regret. This most likely occurs because longer time horizons allow the average cumulative reward from each choice of expert to better approximate the average steady-state reward of that expert. In Fig. \ref{fig:reward}, however, from a cumulative reward standpoint, longer initial time horizons lead to a much slower rate of convergence to the optimal average cumulative reward. This is due to the fact that querying the wrong expert results in a higher corresponding loss of reward at each time step.

We again note that the iteration step of the algorithm is an artificial time-scale imposed over the problem to be able to formulate the problem as an MAB problem. Thus, the trend that a longer time horizon has lower regret after a fixed number of algorithm iterations is not necessarily a valid comparison, since correspondingly more time has elapsed in terms of the MDP. Indeed, given the results in Fig. \ref{fig:results}, and with our stated goal of maximizing the cumulative reward, we would, perhaps counter-intuitively, choose the initial time horizon $T_0=4$ despite the fact that it incurs a larger regret in terms of the algorithm. We note however, that lower time horizons inherently give samples with higher variance, meaning that this trend may not hold true for all MDPs. Further, we note that choosing a larger $T_0$ is not asymptotically penalized since it does, eventually, achieve the same average cumulative reward as the lower $T_0$. Finally we note that all choices of $T_0$ have logarithmically growing regret and the average cumulative reward tends to the average steady-state reward of the best expert in the set.



\begin{figure}[t]
 \centering
 \includegraphics[width=0.75\columnwidth]{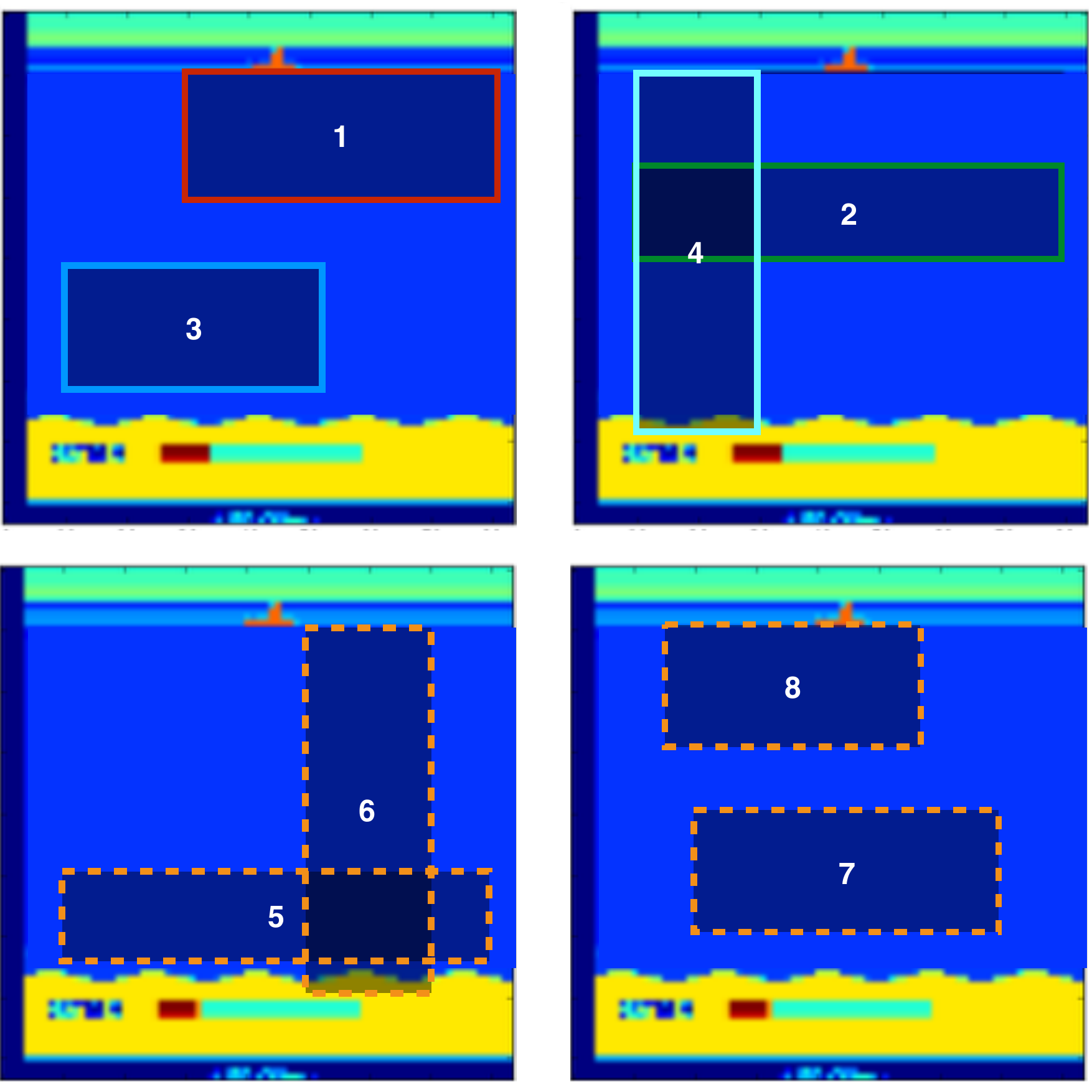}
 \caption{The first row displays occlusions 1 through 4 used for training. Each of the four Q-networks experts was trained against one occlusion, which we denote with the colored borders. The second row displays the new set of occlusions 5 through 8 used for testing, which we denote with the dashed border.} 
\label{fig:occlusions}
\end{figure}

\subsection{High-dimensional Atari Game}
\label{sec:result}
For our second set of experiments we use the Seaquest domain from OpenAI gym \cite{brockman2016openai} shown in Fig. \ref{fig:games_snapshot}. Given the high-dimensional observations of this environment, we chose our experts to be a set of Q-networks as defined in \cite{Minh_2013}. These action-value networks are functions which take a sequence of images as inputs and output a vector of length $|\mathcal{A}|$ corresponding to the value of taking each action. In this work, each Q-network was trained under a fixed occlusion from our set (occlusions 1 though 4, see Fig. \ref{fig:occlusions}) using the algorithm presented in \cite{Minh_2013}. In total, four experts were trained, each against a unique occlusion. After training each expert, the UCB algorithm can be used in conjunction with \textit{any} occlusion to minimize the regret and select at run-time the Q-network in our set of experts which performs best overall.

All high-dimensional experiments were performed on a MacBook Pro 2013 with 2 GHz Intel Core i7 processors. For simplicity, the time horizon $T_n$ for which each expert is executed was held at 100 environmental (MDP) steps and all experiments were performed for a total of 1500 algorithm iterations. With these parameters each run of algorithm \ref{alg:ucb} took approximately 12 minutes. 

\begin{figure}[t]
\centering
\includegraphics[width=0.9\columnwidth]{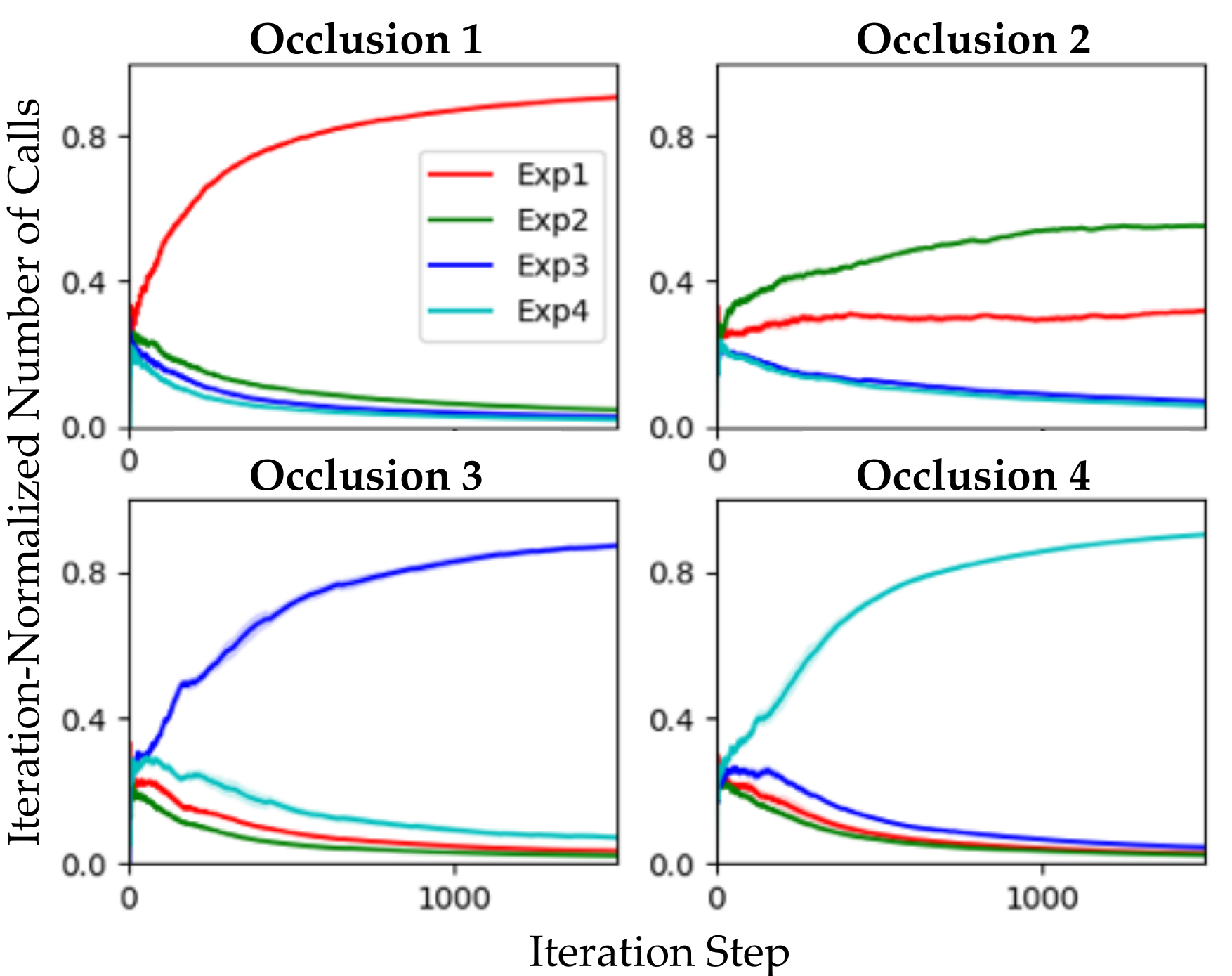}
\caption{For the first four occlusions, each plot shows the iteration-normalized number of calls to each expert against the algorithm's iteration step.} 
\label{fig:alg_perf_k_occl}
\end{figure}

In Sec. \ref{subsub:exp_sel_set_occl}, as a sanity check, we show the performance of the algorithm against the occlusions used at training time. In Sec. \ref{subsub:exp_sel_vs_new_occl}, we run the algorithm against the test occlusions 5 through 8. Finally, in Sec. \ref{subsub:exp_sel_vs_robust}, again using occlusions 5 through 8, we compare the accrued rewards of the algorithm against experts specifically trained to be robust.

\begin{figure*}[t]
\centering
\includegraphics[width=0.85\textwidth]{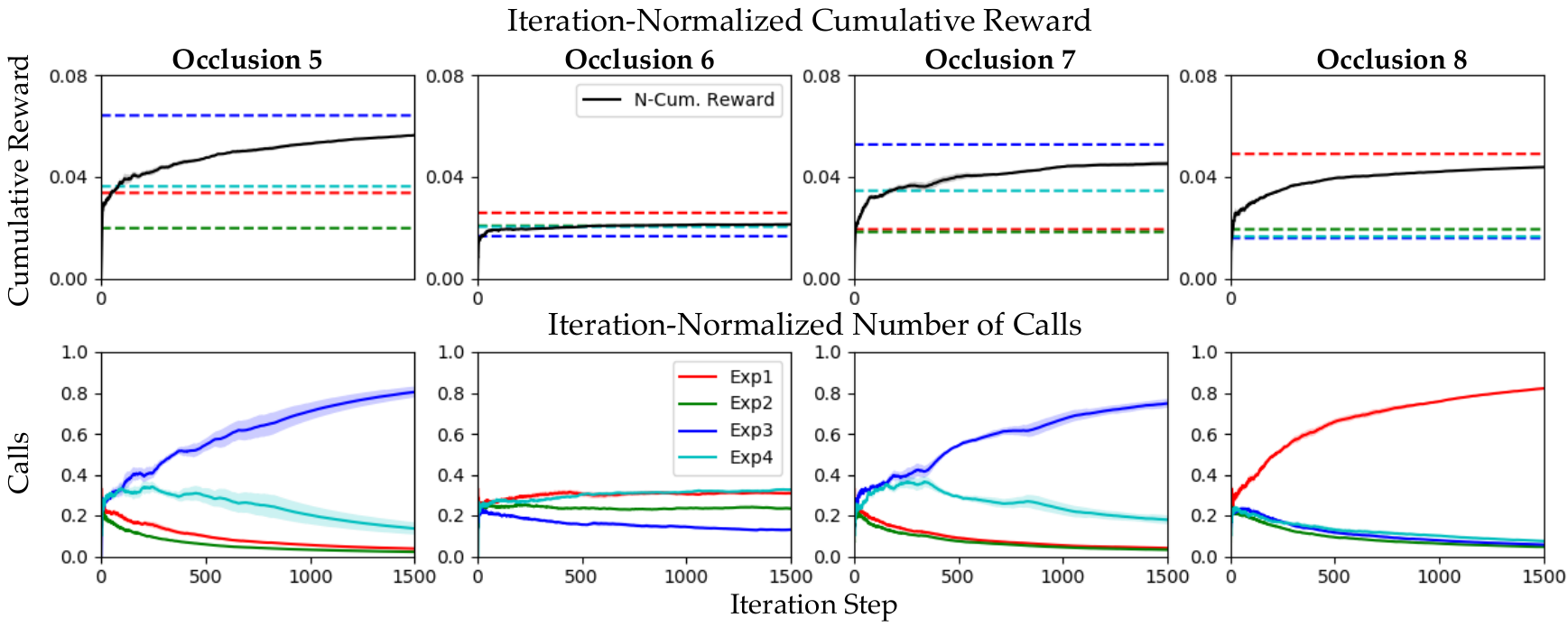}
\caption{Each column shows the average over five runs of the algorithm using a fixed occlusion from the test set. The first row shows the iteration-normalized cumulative reward against the algorithm's time step. The dashed lines show the average accrued 100-step reward for a specific expert. The second row shows the iteration-normalized number of calls to each expert against the algorithm's iteration step.} 
\label{fig:alg_perf_unk_occl}
\end{figure*}

\subsubsection{Expert Selection vs. Training Occlusions}
\label{subsub:exp_sel_set_occl}
~\\
\indent In our experiments, we used four different observation models (in the form of occlusions of the input images) to train each of the experts. The set of occlusions used for training are 1 through 4, shown in Fig. \ref{fig:occlusions}. These were chosen arbitrarily to cover different areas of the input image. Note that in Fig. \ref{fig:games_snapshot} the Seaquest domain looks different than in Fig. \ref{fig:occlusions}: this is due to a preprocessing step on the raw images.



\begin{figure}[!h]
\centering
\includegraphics[width=0.9\columnwidth]{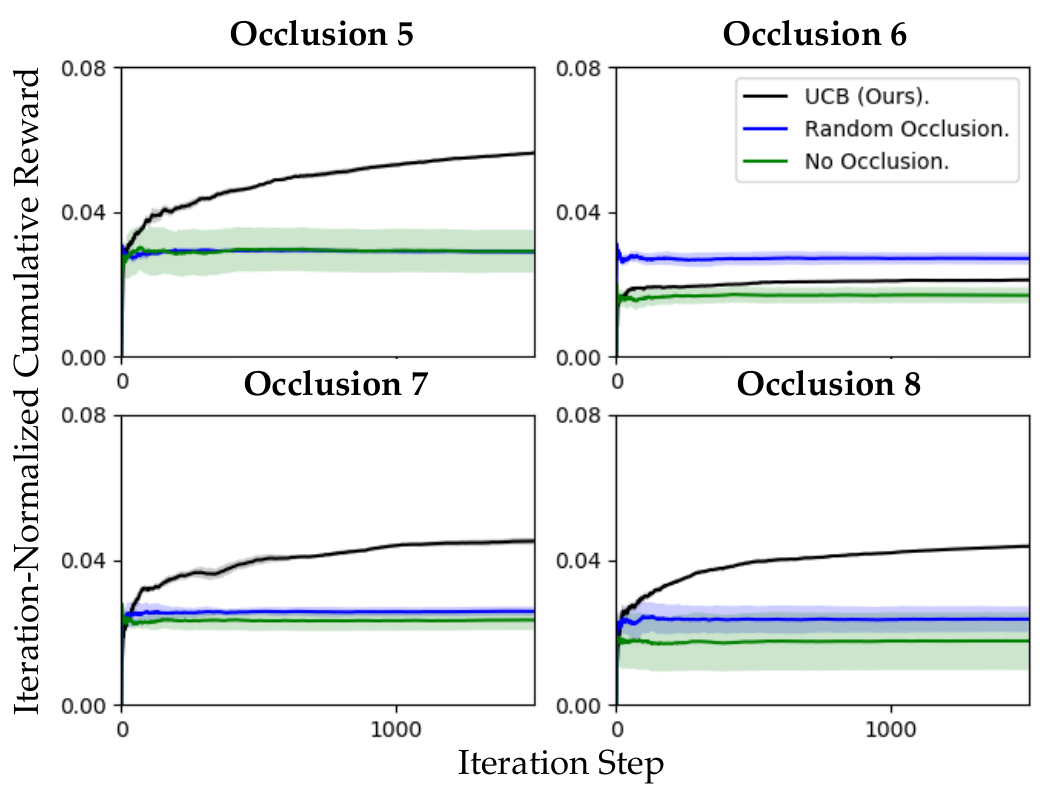}
\caption{Each figure depicts the iteration-normalized cumulative reward for the UCB algorithm, as well as the iteration-normalized cumulative rewards for the 3 Q-networks trained with random occlusions and the 3 Q-networks trained without occlusions.} 
\label{fig:alg_perf_vs_blexps}
\end{figure}

This first test of the algorithm is performed against all the occlusions from our training set. The expected result is that given our set of four experts and one of the occlusions (1 through 4), the UCB algorithm will pick the expert trained against that occlusion most often. Figure \ref{fig:alg_perf_k_occl} shows the performance of the UCB algorithm for the first four types of occlusions. As expected, given an occlusion from our set, the expert that was trained under that type of occlusion performed better than any of the other experts and was picked most often.




It is interesting to note that for the second occlusion the rate at which the correct expert was picked was slower than the rest, despite the fact that the regret normalized by the iteration step did tend towards zero in all cases. This might be caused by a mix of two factors: the generalization capabilities of the Q-networks, as well as the inherent performance differences between experts. While expert 2 was trained with occlusion two as shown in Fig. \ref{fig:occlusions}, expert 1 might have learned to generalize better to other occlusions, thus bringing its 100-step average reward to levels similar to expert 2. The other explanation is that since both experts had full image information for the bottom half of the screen, expert 1's performance might simply be significantly better than expert 2's when the submarine is in the lower half, thus making the 100-step average reward similar for both experts and making it difficult to distinguish which one of the two is best.

\subsubsection{Expert Selection vs. Test Occlusions}
\label{subsub:exp_sel_vs_new_occl}
~\\
\indent A useful feature of this multi-armed bandit algorithm is that it makes decisions solely based on the accrued rewards, and it is therefore agnostic to the input images. In practical terms this means that the types of occlusions we can use at run-time do not need to be those from our training set. The second row in Fig. \ref{fig:occlusions} shows the set of test occlusions.

Figure \ref{fig:alg_perf_unk_occl} shows that for most of the new occlusions, one expert was favored above the rest. For the test occlusions 5 and 7, the algorithm picked expert 3 more often than any other, and for occlusion 8, the algorithm seemed to favor expert 1. An interesting aspect to note is that the experts that were picked more often are also the ones whose occlusions were most similar or had most overlap to the ones they were exposed to at training time. For occlusion 6, the algorithm did not seem to favor any one expert more than the rest within the first 1500 iterations. However, as expected, the two experts that were picked most often were the ones from our set with the highest 100-step average reward denoted with the dashed lines in Fig. \ref{fig:alg_perf_unk_occl}.



\subsubsection{Expert Selection vs. Baseline Experts}
\label{subsub:exp_sel_vs_robust}
~\\
\indent For the last set of high-dimensional experiments we test the algorithm against a three Q-network baselines whose training images included occlusions. At training time, every $10^4$ environmental steps, a new random occlusion is produced to overlay the input images. By including these random occlusions during training, the network becomes more robust. Figure \ref{fig:alg_perf_vs_blexps} confirms this: the Q-networks trained with random occlusions do better on average than Q-networks trained with no occlusions at all. In addition, in three out of the four test occlusions the UCB algorithm outperforms both baselines within 1500 iterations.


Note that the performance of the algorithm is heavily dependent on our set of experts and their ability to generalize. While UCB performed best overall, none of our trained experts had a truly good policy for occlusion 6. The performance could be improved with additional experts trained to deal with occlusions in the right-hand side of the images. In general, having a large and diverse set of experts will perform better than a smaller one. However, having too many experts can also slowdown the algorithm considerably. Finding the right balance between diversity of experts and set size is a very important issue that we will explore in future work.



\section{Conclusion}
\label{sec:conclusion}


In this work we presented a multi-armed bandit algorithm for the purpose of expert selection in (high-dimensional) MDPs. Given a set of expert policies trained under various dynamics models or observation kernels, we show that one can use a variant of the UCB algorithm from the multi-armed bandit literature to choose at run-time the expert that will do best for the current MDP. In our experiments, we test the UCB algorithm on a simple low-dimensional gridworld and a more complicated, high-dimensional Atari game. Despite the high-dimensionality of the game, our algorithm is able to pick the best expert from a set of experts trained under various observation models. Even when the observation model is different from our original one, our algorithm is still able to select the best expert. We finally compare the algorithm to a baseline of experts specifically trained to be robust to occlusions and find that in general the UCB algorithm still outperforms them. 
\printbibliography

\end{document}